\documentclass[journal=aaemcq,manuscript=article, 
]{achemso}
\pdfoutput=1
\usepackage[version=3]{mhchem} 
\usepackage{caption}

\usepackage{graphicx}
\usepackage{amsfonts}
\usepackage{enumitem}
\usepackage{verbatimbox,lipsum}
\usepackage{amssymb}
\usepackage{subfig}
\usepackage{amsmath}
\DeclareMathOperator*{\argmin}{arg\,min}
\usepackage[ruled,vlined]{algorithm2e}
\usepackage{hyperref}

\usepackage[normalem]{ulem}
\useunder{\uline}{\ul}{}

\usepackage[table,xcdraw]{xcolor}
\usepackage{xargs}  

\include{configuration}

\author{Tiago Botari}
\affiliation[usp]{Institute of Mathematics and Computer Sciences, University of S\~ao Paulo, S\~ao Carlos - SP, Brazil}
\email{tiagobotari@gmail.com}
\author{Frederik Hvilshøj}
\affiliation[AU]{Department of Computer Science, Aarhus University}
\email{fhvilshoj@cs.au.dk}
\author{Rafael Izbicki}
\affiliation[ufscar]
{Federal University of S\~ao Carlos}
\email{rafaelizbicki@gmail.com}
\author{Andre C. P. L. F. de Carvalho}
\affiliation[usp]{Institute of Mathematics and Computer Sciences, University of S\~ao Paulo, S\~ao Carlos - SP, Brazil}
\email{andre@icmc.usp.br}

\title{MeLIME: Meaningful Local Explanation for Machine Learning Models}

\begin{document}
\maketitle

\onehalfspacing
\begin{abstract}
Most state-of-the-art machine learning algorithms induce black-box models, preventing their application in many sensitive domains.
Hence, many methodologies for explaining machine learning models have been proposed to address this problem. In this work, we introduce strategies to improve local explanations taking into account the distribution of the data used to train the black-box models. We show that our approach, MeLIME, produces more meaningful explanations compared to other techniques over different ML models, operating on various types of data. 
MeLIME generalizes the LIME method, allowing more flexible perturbation sampling and the use of different local interpretable models. Additionally, we introduce modifications to standard training algorithms of local interpretable models fostering more robust explanations, even allowing the production of counterfactual examples. To show the strengths of the proposed approach, we include experiments on tabular data, images, and text; all showing improved explanations. In particular, MeLIME generated more meaningful explanations on the MNIST dataset than methods such as GuidedBackprop, SmoothGrad, and Layer-wise Relevance Propagation. MeLIME is available on \url{https://github.com/tiagobotari/melime}.
\end{abstract}

\section{Introduction}
\noindent Machine learning (ML) models have been successfully applied to many different application domains, particularly image recognition \cite{szegedy2015going}, natural language processing \cite{xu2015show}, and speech recognition \cite{lecun2015deep}.
Some state-of-the-art ML models even surpass human performance on tasks where machines were previously known to perform poorly \cite{44806}. Despite this success, in many application domains, high predictive power is not the only feature necessary to comply with user expectations, such as healthcare applications \cite{litjens2017survey}. 

In practice, one of the main concerns in ML applications is the black-box nature of the induced models. Such models contain up to hundreds of billions of internal parameters \cite{brown2020language}, representing a very complex computational structure, which makes predictions performed by these models challenging to understand by humans \cite{44806}. In turn, this decreases one's trust in the model, as it is hard to judge the quality of the predictions performed. Thus, providing explanations for the predictions is essential, especially in domains that can significantly impact people's lives, such as medical diagnostics \cite{caruana2015intelligible, tjoa2019survey} and autonomous driving \cite{bojarski2016end}.  
Some areas even require models to be interpretable to allow agencies to check whether the models are in agreement with regulatory laws \cite{goodman2017european}.
Furthermore, explainable models also  facilitate the use of human feedback
on decision processes \cite{human_in_the_loop}.

\begin{figure}[t]
    \centering
    \includegraphics[width=0.8\linewidth]{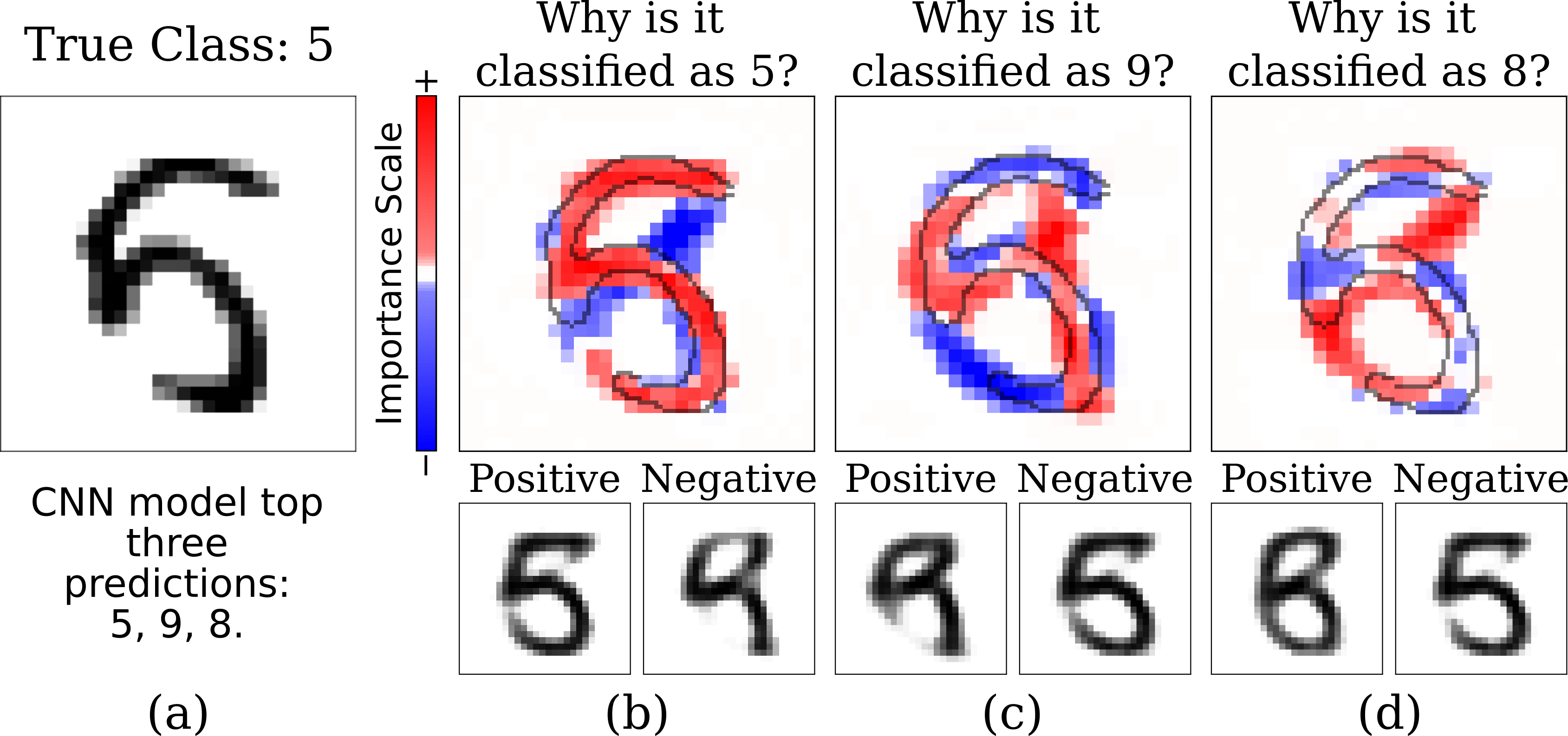}
    \caption{MeLIME local explanations produced for a CNN model trained over the MNIST dataset. 
    An image (number five) and the produced explanations with counterfactual examples for the top three predictions: (a) number five; (b) number six; and (c) number eight.}
    \label{fig:vae_mnist}
\end{figure}

In order to address such demands, a large number of novel methods for explaining ML models have been proposed in recent years \cite{ gilpin2018explaining,molnar2019,binder2016layer,lime, samek2019explainable, samek2020toward, guidotti2019investigating, 8920138}.  
These methods operate on distinct levels, including (i) on the dataset \cite{visualizingFeatures}, (ii) during the design of ML models \cite{selfexplnn,coscrato2019nls}, and (iii) post hoc, after a model has been fitted to a dataset \cite{lime}. In this work, we focus on the post hoc level. We assume that a ML model has already been trained, and we wish to provide explanations for predictions made by this model. In particular, we develop methods applicable to \emph{all} black-box ML models and produce explanations specific to each prediction of interest, known respectively as \emph{model-agnostic} and \emph{local} explanations.

Due to their broad applicability, model-agnostic explanations are promising methodologies, and many such methods have been investigated \cite{friedman2001greedy,fisher2018all,lime,lundberg2016unexpected,vstrumbelj2014explaining}. 
Possibly the most popular method is LIME (Local Interpretable Model-agnostic Explanations) \cite{lime}, which can be applied to classification and regression models. 
However, LIME has shown some weakness in producing reliable explanations for some problems  \cite{alvarez2018robustness, molnar2019}.  
The main weaknesses are related to the following questions:
(i) How to correctly generate sample points on the neighborhood of an instance;
(ii) How can the trade-off between the accuracy of the interpretation model and its interpretability be controlled; (iii) How to choose the number of generated sample points around the instance to be explained.

Recently, explanations from LIME were improved by taking such concerns into account.\cite{10.1007/978-3-030-43823-4_21}
Specifically, the geometry of the data space in terms of $\alpha$-shape was used to estimate the domain of the training data and disallow samples from outside such hulls.
$\alpha$-shape however does not scale to high-dimension spaces, thus such solution only works for low dimensional feature spaces.

In this work, we introduce Meaningful Local Interpretable Model-agnostic Explanations (MeLIME), which
produces meaningful explanations for black-box ML models considering the previous weaknesses pointed out. As an example,
Figure \ref{fig:vae_mnist} shows the MeLIME explanations for MNIST as well as counterfactual examples created by our procedure  (see Section \ref{sec:mnist} for details.)
Section \ref{sec:related} reviews related work. Section \ref{sec:method} introduces MeLIME and shows how it can be used for  different tasks, including regression and classification problems on tabular, images, and text data.
Section \ref{sec:results} 
shows examples of explanations created by MeLIME
on several datasets with different characteristics, as well as comparisons with LIME.
Finally, Section \ref{sec:conclusions} concludes the paper.

\section{Related Work}
\label{sec:related}
In recent years, a significant number of explanation methods for ML models have been developed.
Traditionally, such methods have been categorized by their insights into the model which is being explained (model agnostic or not), whether they produce explanations for the model as a whole (global) or per input (local), and if they produce explanations by example or by feature relevances.
For a complete overview, we refer the reader to the summary written by  \citet{arrieta2020explainable}. 
In this work, we focus on local model agnostic explanations that produce feature relevances: given a black-box ML model and a particular input, the task is to identify which input features are relevant for the prediction of the model.
Other methods in this category include LIME \cite{lime}, gradient-based saliency technique \cite{fong2017interpretable}, Grad-cam \cite{selvaraju2017grad}, Shapley values \cite{shapley}, and SmoothGrad \cite{smilkov2017smoothgrad}, among others.
As our method extends upon LIME, we provide next a brief description of LIME.

\subsection{Local Interpretable Model-agnostic Explanations (LIME)} 

LIME is a model-agnostic method able to produce local explanations for ML models. 
It introduced a general framework to generate a local explanation for any ML model. LIME works as follows.
Given an ML model, $f$, a local explanation can be created for an instance, $x^*$, using an interpretable model, $g \in G$ where $G$ is a set of interpretable models. 
The local model, $g$ is found by minimizing
\begin{equation}
    \xi(x^*) = \argmin_{g\in G} \mathcal{L}(f, g, \pi_{x^*}) + \Omega(g)
\end{equation}
where $\mathcal{L}(f, g, \pi_{x^*})$ is a measure of how unfaithful $g$ is in approximating $f$ over $\pi_{x^*}$ (a measure of locality around $x^*$), and $\Omega(g)$ is a measure of complexity of the local model $g$. 

In practice, the original implementation of LIME uses
$\mathcal{L}$ 
to be a 
squared loss weighted by $\pi_{x^*}$, where 
$\pi_{x^*}$ is an Gaussian kernel and the sum is done over samples  drawn uniformly on the feature space. 
Moreover,
it uses $G$ as a class of linear models and 
$\Omega$ 
is chosen so as to enforce that at most $K$ parameters are selected on the linear model. The interpretation made by LIME does not need to be made on the same space used to construct $g$; a more interpretable feature space may be used.

Variants of LIME have been proposed in the literature, among them K-LIME   \cite{hall2017machine}, LIME-SUP \cite{hu2018locally} and NormLIME \cite{ahern2019normlime}. K-LIME uses clustering techniques (K-means) to partition a dataset into K clusters; each data partition is used to train a local generalized linear model. The K value is tuned to maximize $R^2$ for all local models. LIME-SUP uses a similar strategy; however, it implements a supervised partitioning tree that can improve the produced explanations.  NormLIME aggregate and normalize many local explanations to produce a class-specific global interpretation.

\section{Methodology}
\label{sec:method}
This section describes MeLIME, our strategy to generate meaningful local explanations for machine learning models. Our approach follows a similar framework as LIME  in that 
local interpretable models are used to provide explanations for ML model predictions. However, MeLIME  yields more robust and interpretable results. We considered three key components in designing MeLIME:
(i) a mechanism for generating data around a meaningful neighborhood of $x^*$ that, contrary to LIME, takes into account the distribution of the data (Section \ref{sec:local-models}) , (ii) an interpretable model used to produce local explanations (Section \ref{sec:local-models}), and (iii) a strategy for defining the number of local samples needed in order to obtain a robust local explanation (Section \ref{sec:robustness}).
The strategy that we propose here considers the domain of the data used to induce the ML model, as described in subsection \ref{sec:Feature_space}. 

Our approach for explaining the prediction for $x^* \in \mathcal{X}$ made by a black-box ML
model $f$ that has been fitted using a dataset $D = \{(x_1,y_1),\ldots,(x_n,y_n)\}$ is as follows.
First, we generate samples on a neighborhood of $x^*$ using a generator $\mathbb{G}(x^*,r)$ that encodes what a meaningful neighborhood in $\mathcal{X}$ is.
 This generator can be chosen according to the application; in Section \ref{sec:local-models} we provide several examples 
 that are appropriate for tabular data, images and text data.
Let $(x_1,f(x_1)),\ldots,(x_b,f(x_b))$ be the generated data along with the predictions made by the black-box ML model.
We then 
transform the features $x \in \mathcal{X}$ to a new space $\mathcal{X}'$, $T: \mathcal{X}\rightarrow \mathcal{X}'$, in which it is easier for humans to understand explanations. The user should define the transformation $T$, which will depend on the nature of the desired explanation and the specific knowledge domain of the task. For instance,
while a classifier for text data may be trained using word embeddings, it is often easier to interpret it using a bag-of-words feature space.
When the user wants to use original features, $T$ should be set as an identity transform.   
Let $(x'_1,f(x_1)),\ldots,(x'_b,f(x_b))$ be the resulting dataset. Finally, we fit an interpretable prediction model, $g$, for this set.
During the training process, as $\mathbb{G}(x^*,r)$ encodes a meaningful neighborhood, the locally generated samples can also be
used as counterfactual examples to complement the explanation produced. MeLIME collects the top five favorable and unfavorable samples according to the black-box prediction.
This procedure is repeated until the convergence criteria for the obtained explanation are met (Section  \ref{sec:robustness}).
Algorithm \ref{alg:meLime} summarizes MeLIME. The full implementation of MeLIME is available on \url{https://github.com/tiagobotari/melime}.

\begin{algorithm}[h]
\SetKwInOut{Input}{Input
}
\SetKwInOut{Output}{Output}
\Input{Black-box ML Model, $f$; 
  instance $x^*$ to be explained;
  transformation $T :\mathcal{X} \longrightarrow \mathcal{X}'$ that maps the  original feature space to the space that will be used to produce the local model $g$; generator of samples on a neighborhood of size $r$ around $x^*$,
  $\mathbb{G}(x^*;r)$; batch size $b$; convergence parameters $\epsilon_c$ and $\sigma$}
  \Output{Explanation about why $f(x^*)$ is the prediction for $x^*$}
$\mathcal{D} \gets \emptyset$\;
\Repeat{$\epsilon > \epsilon_c$ and $\delta > \sigma$}
 {
 \For{$i=1,\ldots,b$}{
  $x_i \gets \mathbb{G}(x^*;r)$\;
  $x'_i \gets T(x_i)$\;
  $\mathcal{D} \gets \mathcal{D} \cup \{(x'_i,f(x_i))\} $\;
  }
  train $g$ using $\mathcal{D}$\;
  compute $\epsilon$ and $\delta$, the $g$ training error, and the converge criteria is
  defined in section \ref{sec:robustness}.
 }
 {}
 Get explanations $\alpha$ from $g$ (Section \ref{sec:robustness})\;
 \textbf{return} $\alpha$ \;
 \caption{Strategy to Generate Meaningful Explanation - MeLIME}
 \label{alg:meLime}
\end{algorithm}

\subsection{Generating samples on a meaningful neighborhood}
\label{sec:Feature_space}

To produce a meaningful local explanation, it is important to correctly sample data around $x^*$.
More specifically, the local data (used to fit the interpretable model) needs to be sampled from an estimator that resembles the distribution that generated the original training data.
If this is not the case, the interpretable model may be trained on regions of the feature space that were not used for training the black-box model, potentially leading to inaccurate explanations.

Let $r$ be a parameter that controls the size of the neighborhood. 
We denote by $\mathbb{G}(x^*;r)$ the function that generates a sample point in a neighborhood of size $r$ around $x^*$.
We investigate four different generators, $\mathbb{G}$:

\begin{enumerate}
\label{items}
    \item \textbf{KDEGen}: a \textit{kernel density estimator}  \cite{parzen1962estimation} (KDE) with the Gaussian kernel is fitted by choosing a proper bandwidth $h$ of the kernel. Afterwards, the subset $\mathcal{I}$ of points in the original training data within the radius $r$ from $x^*$ is identified, i.e., $\mathcal{I} = \{ x\in \{x_1,\ldots,x_n\} : d(x^*, x) \leq r\}$ where $d$ is a distance function. Local samples are then generated by repeatedly drawing $\tilde{x} \sim \texttt{Uniform}(\mathcal{I})$ to successively sample a new point $x_{\texttt{new}} \sim K(\cdot,\tilde{x})$.  
    For a Guassian kernel,  this is identical to standard sampling from Gaussian KDEs except from the fact that we sample only from training samples within $r$. See the algorithm \ref{alg:KDEGen} for details.
    
    \item \textbf{KDEPCAGen}: 
    KDEGen can be ineffective if the number of features is large. Thus, KDEPCAGen first performs \textit{Principal Component Analysis} (PCA) to map the instances into a lower dimensional space by applying a transformation $W : \mathcal{X} \rightarrow \mathbb{R}^m$, where $m$ is a small number chosen by the user or by setting a minimal value for the cumulative explained variance ratio. After the data is projected to $\mathbb{R}^m$, KDEGen is applied on this space. After generating a new instance $z \in \mathbb{R}^m$, it is mapped back to $X$ by using the approximated inverse transformation $W^{-1}$ (where components with small eigenvalues are zeroed).

    \item \textbf{VAEGen}: First, a \textit{Variational Auto Encoder} (VAE) \cite{kingma2019introduction} is trained over the dataset $D$.
    Let $q_\phi(z|x)$ denote the encoder and $p_\theta(x|z)$ denote the decoder of the VAE, where $z$ are the latent variables. Given a sample $x^*$ and a neighborhood size $r$, the new sample is generated by (i) encoding $x^*$ via $z^* \sim q_\phi(z|x^*)$, (ii) drawing $\epsilon \sim \texttt{Uniform}(-r,r)^m$
    where $m$ is the dimension of $z$, and (iii) decoding $x_{\texttt{new}} \sim p_\theta(x|z^* + \epsilon)$. See the algorithm
    \ref{alg:VAEGen} for the details of this procedure.
    
    \item \textbf{Word2VecGen}: 
    Consider text data as a set of tokens $\mathbb{T}$.
    We first train word2vec \cite{mikolov2013distributed} embeddings for $\mathbb{T}$ using the training corpus.
    Let $\psi(t) \in \mathbb{R}^m$ be the word2vec representation of token $t \in \mathbb{T}$. In order to sample a new instance around $x^*$, we first extract its tokens. Let $t(x^*)$ be the vector that contains such tokens. 
    We then choose one element of $t(x^*)$ at random; say $t_i$.
    Finally, we replace  $t_i$ in $t(x^*)$ by a neighbor token drawn at random from the set $\{t \in \mathbb{T}: d(\psi(t),\psi(t_i))\leq r\}$. See the algorithm
    \ref{alg:Word2VecGen} for the details.
\end{enumerate}

\begin{algorithm}[h]
\SetKwInOut{Input}{Input
}
\SetKwInOut{Output}{Output}
\Input{Instance $x^*$;   neighborhood size $r$ around $x^*$,
smoothing kernel
 $K(x,x^*)$,  samples  $D_x=\{x_1,\ldots,x_n\}$}
  \Output{new sample point $x_{\text{new}}$}
Let $\mathcal{I} = \{ x\in D : d(x^*, x) \leq r\}$

Sample $ \tilde{x} \sim \text{Uniform}(\mathcal{I})$\;
Sample $x_{\texttt{new}} \sim K(\cdot,\tilde{x})$\;
 \textbf{return} $x_{\text{new}}$ \;
 \caption{KDEGen}
 \label{alg:KDEGen}
\end{algorithm}

\begin{algorithm}[h]
\SetKwInOut{Input}{Input
}
\SetKwInOut{Output}{Output}
\Input{Instance $x^*$; Encoder: $q_\phi(z|x)$;  Decoder: $p_\theta(x|z)$; neighborhood size $r$ around the representation of $x^*$ in the latent variable, $z^*$.}
  \Output{new sample point $x_{\text{new}}$}
$z^*  \sim q_\phi(z|x^*)$\;
$\epsilon \sim \texttt{Uniform}([-r,r])^m$\;
$z \xleftarrow{} z^* + \epsilon$\;
$x_{\text{new}} \sim p_\theta(x|z)$\;
\textbf{return} $x_{\text{new}}$ \;
\caption{VAEGen}
\label{alg:VAEGen}
\end{algorithm}

\begin{algorithm}[H]
\SetKwInOut{Input}{Input
}
\SetKwInOut{Output}{Output}
\Input{Set of tokens $\mathcal{S}^*$ of sentence $x^*$; a token corpus $s \in \mathcal{S}$; neighborhood size $r$; $Encoder: \mathcal{S}\rightarrow Z$}
  \Output{new sample point $x_{\text{new}}$}
 $s_j \sim Uniform(\mathcal{S}^*)$ \;
 $z^*_j \leftarrow Encoder(s^*_j)$ \;
 $z_i \leftarrow Encoder(s_i), \quad \forall s_i \in \mathcal{S}$ \;
 $\mathcal{I} \leftarrow \{s_i : d(z_i, z^*_j) \leq r\}$\;
$x_k \sim Uniform(\mathcal{I})$\;
 Generate $x_{\text{new}}$ replacing the token $s_j$ by $s_k$ in $S^*$\;
 \textbf{return} $x_{\text{new}}$ \;
 \caption{Word2VecGen}
\label{alg:Word2VecGen}
\end{algorithm}

\subsection{Interpretable Local Models}
\label{sec:local-models}

After a local dataset has been generated using the methods from Section \ref{sec:Feature_space},
we fit a local interpretable model to it. 
We implement the following models: 

\begin{enumerate}
    \item \textbf{Local Linear Model}: Using a linear model, we can obtain the explanation by analyzing the angular coefficients of the model \cite{lime}. 
    
    \item \textbf{Local Regression Tree Model}: The decision rules of the tree, together with the feature importance, are used to generate an explanation \cite{freitas2014comprehensible}.
    
    \item \textbf{Local Statistical Measures}: We can produce the explanation by extracting simple statistical measures from the ML model's prediction over the local perturbations.  More precisely, let $x^*=(x_1, \ldots, x_d)$ be the instance to be explained. Afterwards, we compute summary statistics about the predictions obtained when creating perturbations over each dimension $x_i$. In our experiment, we chose to use the mean, median, standard deviation measures. 
\end{enumerate}

Other models that are known to be easy to interpret could also be used.  We designed MeLIME to allow easy inclusion and configuration of interpretable local models.
For other possible interpretable models, we refer the reader to the following references \cite{molnar2019, arrieta2020explainable}.

\subsection{Robustness of Explanations}
\label{sec:robustness}

The local model fit depends on how many instances are generated around $x^*$. 
In order to ensure the convergence and stability of the local model $g$, we 
implement a local-mini-batch strategy.
That is, we generate local instances while training the local model $g$ until it converges according to the specific criteria presented below. 
The criteria encourages explanations that are robust. 
As summary statistics for the explanations, we use the coefficients of the model, the feature importances, and the feature means for the Local Linear Model, the Local Regression Tree model, and the Local Statistical Measures, respectively.

Let $\alpha(g) \in A$ be the set of summary statistics (explanations) given by the  local model $g \in G$. 
Our mini-batch procedure works as follows. 
First, we create a batch of instances of size $b$ around $x^*$ and fit our local model to it. 
Let $g_1$ be the fitted model. 
We then compute $\alpha_1:=\alpha(g_1)$. 
Next, we create $b$ additional instances and refit our local model using the whole set of instances. 
Let $g_2$ be the fitted model, $\alpha_2:=\alpha(g_2)$, and

\begin{equation}
\delta = \frac{1}{\texttt{dim}( A )}|| \alpha_2 - \alpha_1||_1. 
\end{equation}
If $\delta<\sigma$, where $\sigma$ is a previously specified value,
the convergence of the feature importance is achieved.
 Otherwise, we repeat this procedure until $\delta \leq \sigma$. 

Moreover, we also used the convergence of the local model's fitting error, $\epsilon$, as a convergence criterion. The local model will be trained using newly generated local-mini-batch until the achievement convergence criteria for $\epsilon$ and $\delta$.

\section{Experimental Evaluation}
\label{sec:results}
To assess the strategies' performance in producing meaningful local explanations, we performed different experiments for various classes of ML models in distinct application domains. We investigated ML tasks of regression and classification. As a regression ML task, we used a toy model - Spiral Length - where the model predicts a spiral length. For classification tasks, we have chosen classification of tabular data using the Iris dataset \cite{fisher1936use}, image using the MNIST dataset \cite{lecun1998gradient}, and sentiment analysis on movie reviews \cite{text-rotten-tomatoes, Pang+Lee:05a}. The codes of the performed experiments are available on \url{https://github.com/tiagobotari/melime}.

\begin{figure*}[t]%
    \centering
    \includegraphics[width=0.9\linewidth]{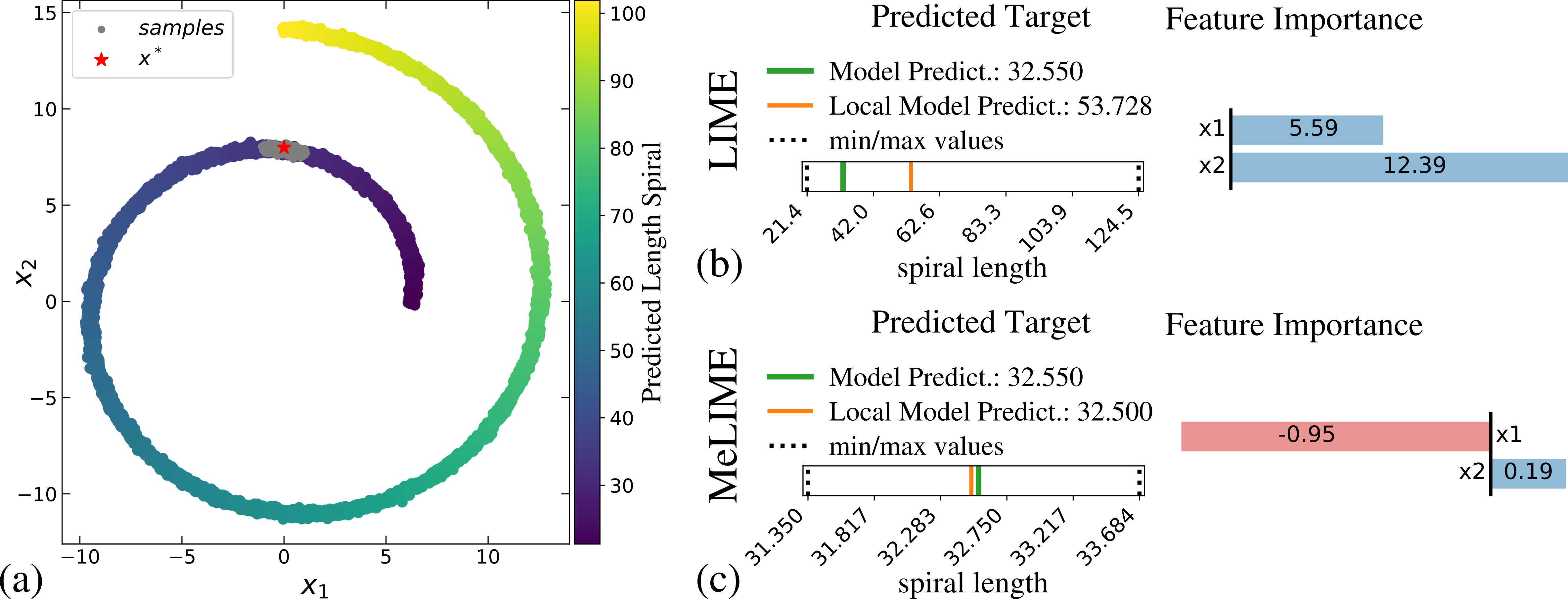}
    \caption{(a) Data generated from the \textit{toy model}  - Length of Spiral, Equation \ref{eq:spiral}. The color scale corresponds to the predicted values of the MPL model. The red star represents a data point $x^*=(0, 8)$, and the gray points the sampled data from KDEGen. Explanations produced for an instance $x^*$ using (b) LIME; and (c) MeLIME methodology with KDEGen and a linear model as local model.}
    \label{fig:spiral}
\end{figure*}

\subsection{Toy Model: Spiral Length}

The Spiral Length is a Toy Model where the ML task is to predict the Length of a Spiral 
in a 2D space.
The equation that produces the spiral is given by
\begin{eqnarray}
    x_1 &=& \theta \cos(\theta) + \epsilon_1 ~~~~~~ x_2 = \theta \sin(\theta) + \epsilon_2 \\
    y &=& \frac{1}{2}\left[ \theta \sqrt{1+\theta^2}+ \sinh^{-1} \theta \right] \nonumber
    \label{eq:spiral}   
\end{eqnarray}
where $x = (x_1, x_2)$ is a point in the Cartesian plane defined by the spiral representing the features $x_1$ and $x_2$, $\theta$ is an independent variable which decides the ``position'' on the spiral, $\epsilon_1, \epsilon_2 \sim \mathcal{N}(0, 0.1)$ is random noise, and the target value is given by $y$, which is the length of the spiral calculated at a point $x$.

Using Equation \ref{eq:spiral}, we generated 10k data points. 
We divided the data into two sets, training with 80\% and test with 20\%.
We used the training set to train a Multilayer Perceptron (MLP) network implemented in the scikit-learn package \cite{scikit-learn}. 
We then tested the model over the test set, obtaining  $R^2=0.999$ and a mean square error of $0.011$.

We  expect   the spiral's length around $x^*=(0, 8)$ to be  highly dependent on the variable $x_1$ and a small dependence on variable $x_2$. 
Thus, a good local explanation methodology would capture that $x_1$ is the most important feature and a decrease of $x_1$ value would increase y (length of the spiral). 

To produce local explanations, we used LIME and MeLIME. For MeLIME, we used the KDEGen strategy to generate the local-mini-batches and a linear model as local model. Figure \ref{fig:spiral} shows a comparison of the interpretations produced for $x^*$ by MeLIME and that produced by LIME.
LIME provides an explanation that gives high importance for $x_2$, which contradicts our expectations. 
On the other hand, MeLIME produces a local explanation that matches the expected and gives high importance to variable $x_1$.  
Additionally, the MeLIME explanation indicates the correct tendency of for small changes of the features. An increase in $x_1$ variable will cause a decrease in the spiral length. This is indeed the case in the neighborhood of $x^{*}$ according to Equation \ref{eq:spiral} and so with the MLP model.  
 
Analyzing the explanation produced by LIME, we can verify that the incorrect feature importance is related to sample instances belonging to regions out of the spiral domain. The perturbations are generated from a Gaussian that wrongly samples data from regions where the spiral is not defined, which will produce no meaningful prediction from the original ML model.

\subsection{Classification Problem: Iris Dataset}

\begin{figure}[t]
\centering
\includegraphics[width=0.8\linewidth]{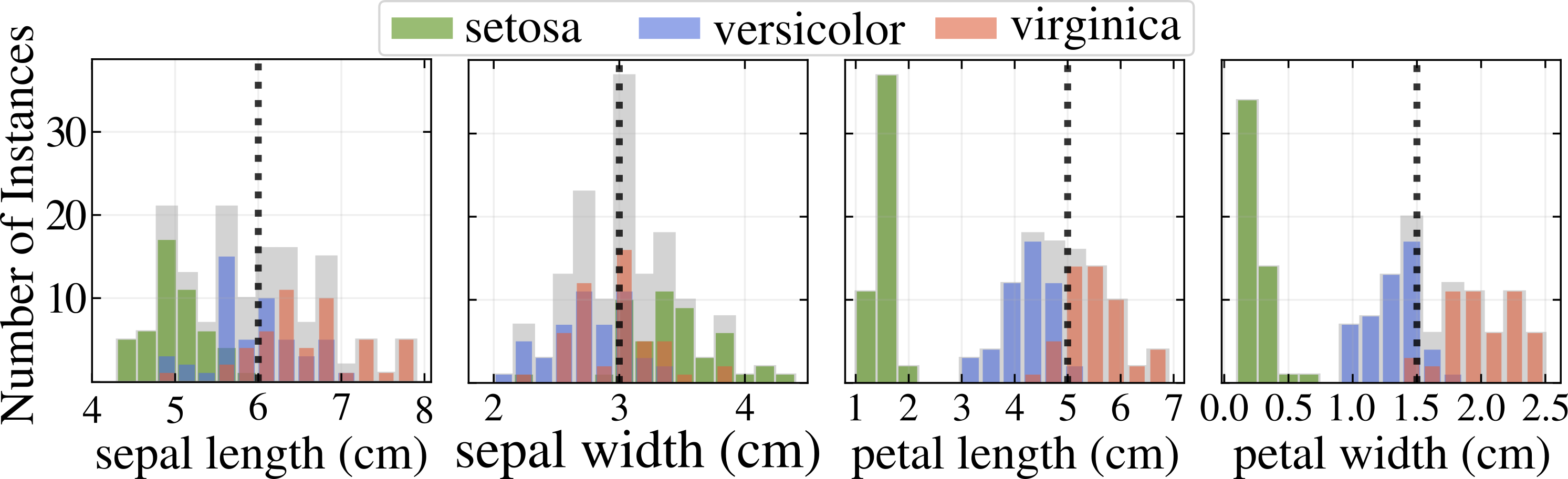}
\caption{Iris dataset distribution over the four features and the three classes. The red line represent the instance $x^*=(6.0, 3.0, 5.0, 1.5)$. The pairwise relationships for the Iris dataset is on the supplementary material. }
\label{fig:iris}
\end{figure}

\begin{figure}[t]
\centering
\includegraphics[width=0.6\linewidth]{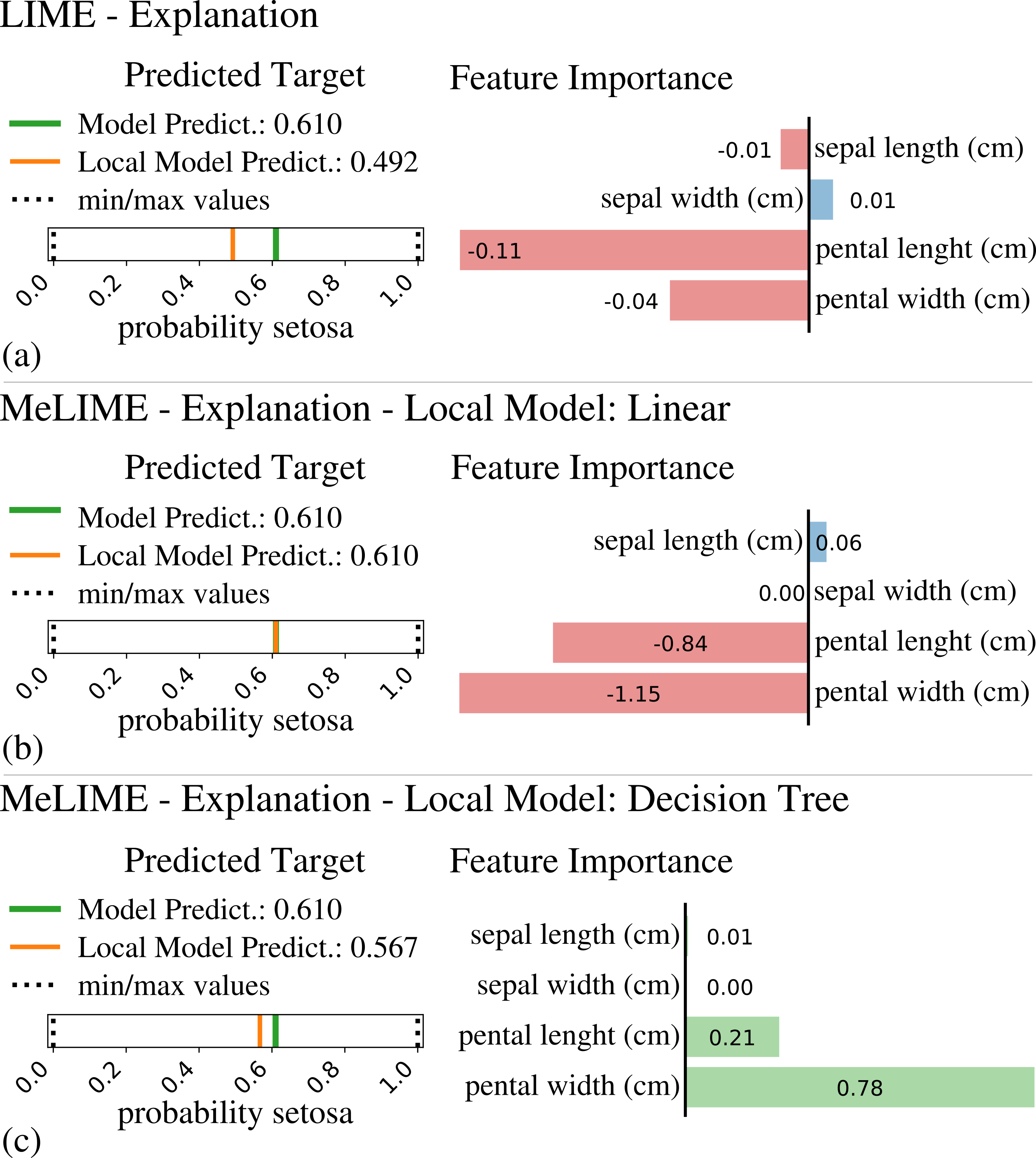}
\caption{Local explanation for an instance $x^*=(6.0, 3.0, 5.0, 1.5)$ produced by an RF model trained on the Iris dataset. (a) LIME explanation; (b) MeLIME using KDEPCAGen and a linear model as local model; (c) MeLIME using KDEPCAGen and a decision tree as local model (decision tree plot is shown in figure  \ref{fig:iris_explantion_tree}).
}
\label{fig:density_lime}
\end{figure}

\begin{figure}[h]
\centering
\includegraphics[width=1.0\linewidth]{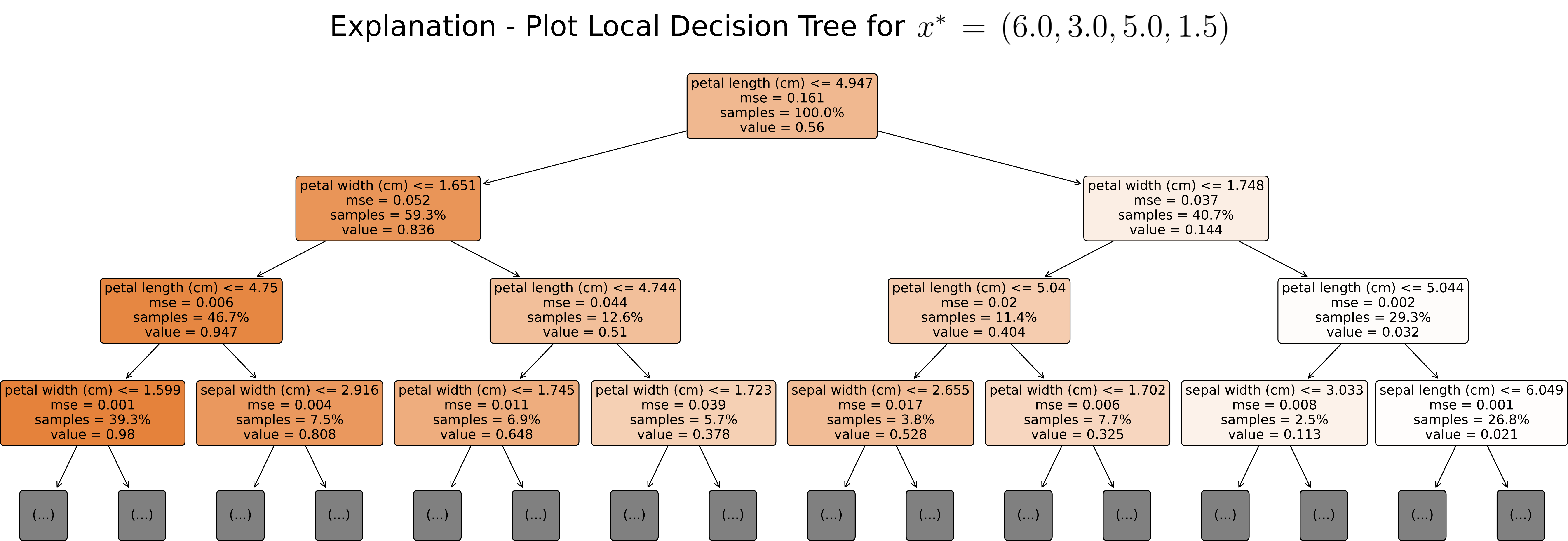}
\caption{Explanation using a Decision Tree as local model. The ML model was trained using the Iris dataset. The explanations was produced using MeLIME instance $x^*=(6.0, 3.0, 5.0, 1.5)$. }
\label{fig:iris_explantion_tree}
\end{figure}

In this subsection, we analyze the explanations produced for a ML model trained on the Iris Dataset \cite{fisher1936use}. The task is to classify instances of Iris flowers into three species: \textit{setosa}, \textit{versicolor}, and \textit{virginica}. 
The dataset has 150 instances with four features representing the sepal and petal width and length in centimeters. 
We represent an instance as $x=(\textrm{sepal-length}, \textrm{sepal-width}, \textrm{pental-length}, \textrm{pental-width})$. A visualization of the dataset is shown in Figure \ref{fig:iris}.

To perform our experiments, we split the dataset into two sets, training and testing. We randomly selected $80\%$ and $20\%$ of the instances for the training and test sets. Using the training set, we fitted a Random Forest (RF) classifier implemented in the scikit-learn package \cite{scikit-learn} using the default tuning parameters (see supplementary for details). The fitted model has an accuracy of 0.97 on the test set.

We now analyze the explanations given by both LIME and MeLIME for an instance $x^*=(6.0, 3.0, 5.0, 1.5)$, represented by the dotted vertical lines in Figure \ref{fig:iris}. 
For $x^*$,  the petal length and width values can separate almost all instances of the versicolor and virginica classes (Figure \ref{fig:iris}), and thus 
we  expect a local explanation to point to these quantities as the most import features.
Moreover, an increase in the values of the pental's length and width should decrease the RF model's probability of classifying an instance as iris versicolor.

Figure \ref{fig:density_lime} shows the explanations obtained. We used the strategy KDEPCAGen in MeLIME, and two  local models to produce the feature importance, a linear model and a decision tree (subsection \ref{sec:local-models}).  
The results obtained by MeLIME using both local models are compatible with each other and provided meaningful explanations: up to a sign, both explanations give higher importance to the petal length and width, which agrees with our previous analysis. The explanation produced by LIME is not very discrepant with our expectations. However, the difference between the back-box model and the local model is high (around $0.12$), 
which can decrease one's trust in the explanations given by this method. This is not the case for MeLIME, which has a difference of at most $0.04$. 

\subsection{Classification of Images - MNIST Dataset}
\label{sec:mnist}

The MNIST dataset contains 70k images of handwritten digits \cite{lecun1998gradient}.
The goal is to classify the digits in one of ten classes, labeled
from 0 to 9. 
We use the regular split of the dataset; 60k instances for training and 10k for test. 

Using the training set, we fit a CNN model yielding a 98\% accuracy on the test set.
The architecture of the CNN model is available on the Supplementary Material.

To produce an explanation of an instance for the CNN model, we employed MeLIME using VAEGen and a linear model as the local model. We fit the VAEGen using the training set. Next, we selected a random instance from the testing set and investigated the production of explanations using MeLIME. The chosen image represents the number five (true class). Using the CNN model, we obtain the top three class predictions: number five, number nine, and number eight.

Using MeLIME, we obtained a local explanation for the CNN model prediction as number five for the input image under consideration. 
The obtained explanation is in excellent resolution and provides
a precise categorization of the pixels that contribute  positively and negatively for the CNN model classifying the image as number five, Figure \ref{fig:vae_mnist} (a). Analyzing the positive pixels, we can see the pixels draw a better representation of a number five. On the other hand, the negative pixels present patches that, if filled, would possibly confuse the classifier with the number six, eight, nine, and even four. Additionally, MeLIME gives the counterfactual favorable and contrary to be classified as number five, as shown in the bottom of Figure  \ref{fig:vae_mnist} (a).

\begin{figure*}[t]%
    \centering
    \includegraphics[width=1.0\linewidth]{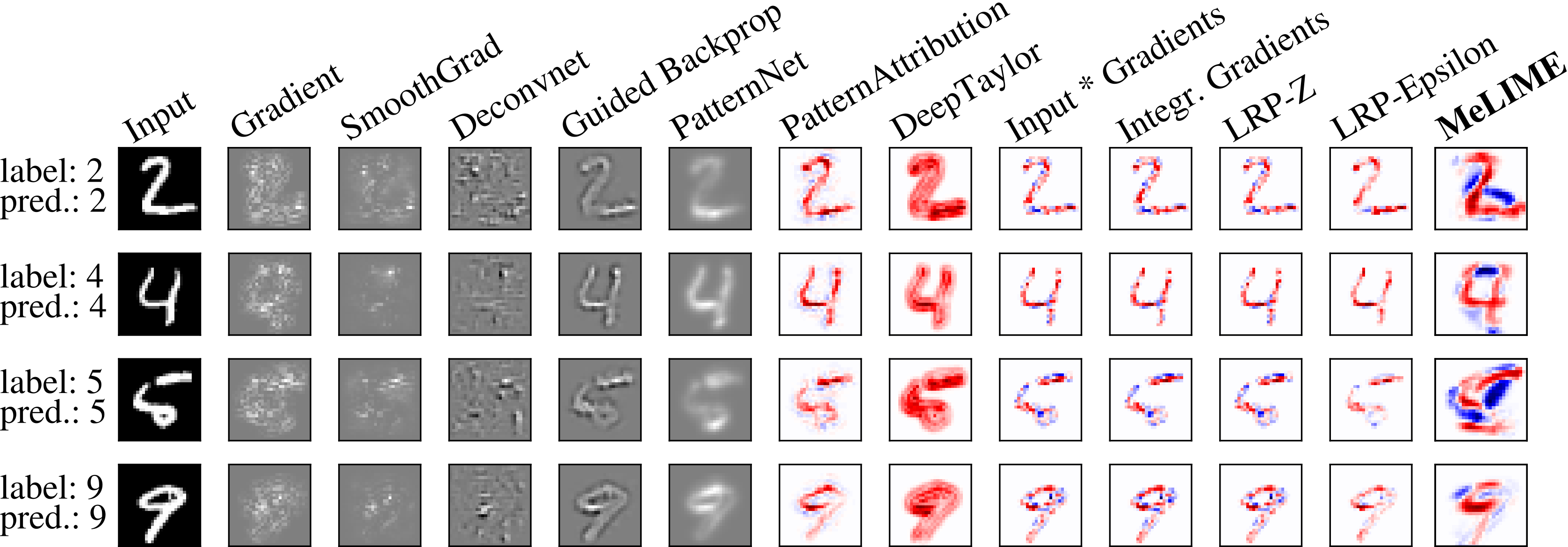}
    \caption{Comparison of explanations produced by different methodologies for a CNN model trained on MNIST dataset.}
    \label{fig:mnist-comparing}
\end{figure*}

Moreover, we investigated possible explanations of why
the CNN model could misclassify
the image as number nine and eight, see Figure \ref{fig:vae_mnist} (b) and (c). We can verify the necessary pixels to modify the classification performed by the CNN model. 
For instance, to make the CNN model more confident that the the input is a nine, it is necessary to close the top loop, as shown in Figure \ref{fig:vae_mnist} (b).
Additionally, MeLIME produced the counterfactual for the possible missclassification classes, which is presented on the bottom of Figure \ref{fig:vae_mnist} (b) and (c).

Finally, we compared MeLIME with other methodologies using the iNNvestigate library \cite{alber2018innvestigate}, as shown  in Figure \ref{fig:mnist-comparing}. We include comparisons with SmoothGrad \cite{smilkov2017smoothgrad}, DeConvNet \cite{Zeiler_2014},  Guided BackProp \cite{springenberg2014striving}, PatternNet \cite{kindermans2017learning}, DeepTaylor \cite{montavon2017explaining}, PatternAttribution \cite{kindermans2017learning}, \cite{bach2015pixel},  IntegratedGradients \cite{sundararajan2017axiomatic}, and Layer-wise Rel-evance Propagation \cite{bach2015pixel}.  The explanation from MeLIME is substantially different in that it allows one to evaluate the negative and positive contribution of a significant set of pixels of the image. This brings additional insights into the black-box classifier.

\subsection{Sentiment Analysis Texts - Movie-Review}

In this section, we present the experiments performed on sentiment-analysis using a movie reviews dataset from the Rotten Tomatoes web site \cite{text-rotten-tomatoes, Pang+Lee:05a}. 
The dataset contains 10662 movie-reviews with the sentiment polarity, i.e., positive or negative. 
Using this dataset, we generated a ML model for sentence classification. 
We first split the data into two sets: training (with $80\%$ of the instances) and test (with $20\%$). 
We vectorized the sentences using a pipeline provided by \verb|CountVectorizer->TfidfTransformer|. 
Using the training set, we trained a Naive Bayes classifier implemented on scikit-learning \cite{scikit-learn}. 
The accuracy of the model over the test set was 75\%.  

\begin{figure*}[h!]
    \centering
    \includegraphics[width=1.0\linewidth]{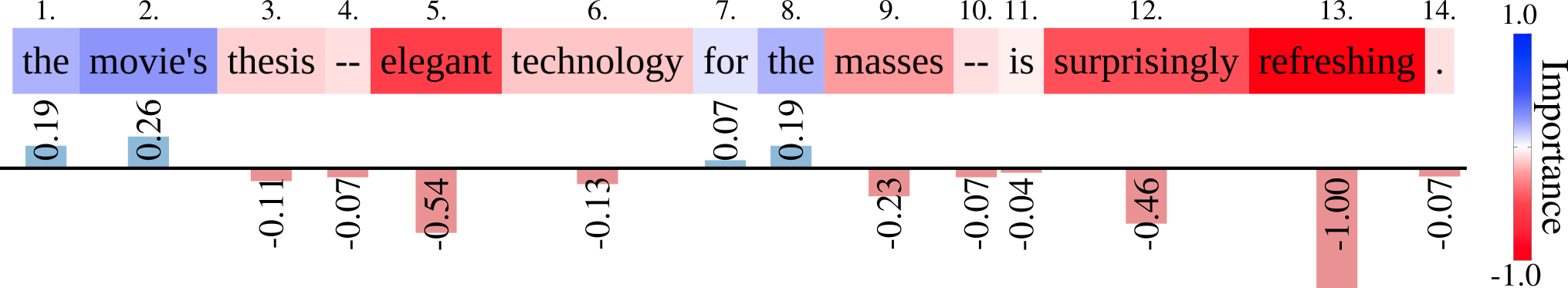}
    \caption{A local explanation was produced using MeLIME for a Naive Bayes classifier model trained on sentences from movie-reviews of the Rotten Tomatoes website \cite{text-rotten-tomatoes, Pang+Lee:05a}. Words/positions showed as reddish color, if changed, are likely to turn the sentence negative while bluish color to become the sentence positive.
    }
    \label{fig:text_importance}
\end{figure*}
\begin{table}[]
\caption{The original phrase to be explained; Favorable and unfavorable phrases are generated for predicting the sentence as positive.}
\label{tab:text}
\resizebox{\textwidth}{!}{%
\begin{tabular}{rlc}
\hline
\hline
\cellcolor[HTML]{FFFC9E}\textbf{N.}&\cellcolor[HTML]{FFFC9E}\textbf{Original phrase} & \cellcolor[HTML]{FFFC9E}\textbf{Prob.}\\
\hline
-&the movie's thesis -- elegant technology for the masses -- is surprisingly refreshing . & 0.710 \\ 
\hline
\hline
\cellcolor[HTML]{adddff}&\cellcolor[HTML]{adddff}\textbf{Favorable phrases}      &                                                 \cellcolor[HTML]{adddff}\\ \hline
1&the movie's thesis -- elegant technology for \textcolor{blue}{\textbf{touching masses}} -- is surprisingly refreshing . & 0.823 \\
2&\textcolor{blue}{\textbf{touching}} movie's thesis -- elegant technology for the masses -- is surprisingly refreshing . & 0.823 \\
3&the \textcolor{blue}{\textbf{touching}} thesis -- elegant technology for the masses -- is surprisingly refreshing . & 0.821 \\
4&the movie's thesis -- elegant technology for \textcolor{blue}{\textbf{wonderful}} masses -- is surprisingly refreshing . & 0.820 \\
5&\textcolor{blue}{\textbf{wonderful}} movie's thesis -- elegant technology for the masses -- is surprisingly refreshing . & 0.820 \\ 
\hline
\hline
\cellcolor[HTML]{ffd9cc}&\cellcolor[HTML]{ffd9cc}\textbf{Unfavorable phrases} &   \cellcolor[HTML]{ffd9cc}                    \\ 
\hline
6&the movie's thesis -- \textcolor{red}{\textbf{ill-conceived}} technology for the masses -- is surprisingly refreshing . & 0.466\\
7&the movie's thesis -- elegant technology for the masses -- is surprisingly \textcolor{red}{\textbf{heavy-handed}} . & 0.468\\
8&the movie's thesis -- elegant technology for the masses -- is surprisingly \textcolor{red}{\textbf{dull}} . & 0.470\\
9&the movie's thesis -- elegant technology for the masses -- is surprisingly \textcolor{red}{\textbf{pretentious}} . & 0.491\\
10&the movie's thesis -- elegant technology for the masses -- is surprisingly \textcolor{red}{\textbf{unfunny}} . & 0.492\\
\hline
\end{tabular}
}
\end{table}


To produce an explanation, we used MeLIME with Word2VecGen to estimate the feature space domain. 
We used the local model strategy \textbf{Local Statistical Measures} (see Section \ref{sec:local-models} for details). 
We selected an instance from the test set and generated an explanation. 
The obtained explanation assigns high importance for the word ``refreshing'' with a negative signal. 
The explanation can be interpreted as if we replace the word ``refreshing'' in the sentence, it will be likely to push the classification towards a negative sentiment. 
Figure \ref{fig:text_importance} shows the explanation produced by MeLIME. 

To further investigate the produced explanation's reliability, we collected artificial sentences generated by Word2VecGen during the local model's training process. We selected the top five sentences classified as positive and the bottom five classified as negative by the ML model, as shown in Table \ref{tab:text}. 
Analyzing the artificial sentences, we can verify that the most negative and positive sentences are generated by replacing the words assigned as most important in the MeLIME explanation, which increases the trust in the generated explanation (Figure \ref{fig:text_importance}).

Analyzing the generated sentences, we would not expect that some of them would increase the likelihood of being classified as positive.  
This is the case for sentence 4 from Table \ref{tab:text}: \textit{``the movie’s thesis – elegant technology for \textbf{wonderful} masses – is surprisingly refreshing.''}. 
Despite that, the ML model returns a high positive score for those sentences, which decreases the trust in the black-box model. 
The ability to allow the user to critically analyze the ML model is a desired property of the explainability methodologies.

\section{Conclusions}
\label{sec:conclusions}
In this work, we show how to produce meaningful local explanations for models induced by different ML algorithms, for distinct application domains.
For such, we 
developed novel strategies that take into account the domain of the feature space used to generate a black-box model. Moreover, we introduced strategies for training the local interpretable model that considers convergence criteria for the feature importance and the local model training error. Using these strategies, we showed that it is possible to obtain a state-of-the-art explanation. 

Our methodology was implemented in a new framework, called MeLIME. Using MeLIME, we produced local explanations for ML models trained on tabular data on regression and classification tasks. We compared MeLIME explanations with that from LIME, analyzing the weakness of LIME that MeLIME solved. 
Moreover, we produced explanations for the MNIST dataset, which allowed us to obtain 
an improvement on the interpretation when compared to other methodologies.
We also performed experiments on sentiment analyses for text data. The produced explanation demonstrated a high capacity for investigating weaknesses of a back-box model. 

We also showed that MeLIME can produce counterfactual examples using the generator  of  local perturbations. The counterfactuals can be used as a complement to the local explanation, thus providing additional insights about the ML model. Furthermore, such artificial samples can be used to discover new instances of interest. 

\section{Acknowledgments}

The authors would like to thank CAPES and CNPq (Brazilian Agencies) for their financial support. T.B. acknowledges support by Grant 2017/06161-7, S\~ao Paulo Research Foundation (FAPESP). R. I. acknowledges support by Grant
2019/11321-9  (FAPESP)
and Grant 306943/2017-4 (CNPq).
The authors acknowledge Grant 2013/07375-0 - CeMEAI - Center for Mathematical Sciences Applied to Industry from S\~ao Paulo Research Foundation (FAPESP).

\bibliography{bib}

\end{document}